# Classification of fetal compromise during labour: signal processing and feature engineering of the cardiotocograph

M. O'Sullivan*[1], T. Gabruseva*[1], GB. Boylan[1,4], M. O'Riordan[1,2], G. Lightbody[1,3], W. Marnane[1,3]

*Abstract*— Cardiotocography (CTG) is the main tool used for fetal monitoring during labour. Interpretation of CTG requires dynamic pattern recognition in real time. It is recognised as a difficult task with high inter- and intra-observer disagreement. Machine learning has provided a viable path towards objective and reliable CTG assessment. In this study, novel CTG features are developed based on clinical expertise and system control theory using an autoregressive moving-average (ARMA) model to characterise the response of the fetal heart rate to contractions. The features are evaluated in a machine learning model to assess their efficacy in identifying fetal compromise. ARMA features ranked amongst the top features for detecting fetal compromise. Additionally, including clinical factors in the machine learning model and pruning data based on a signal quality measure improved the performance of the classifier.

*Keywords—biomedical signal processing, machine learning, features engineering, fetal monitoring, cardiotocography*

## I. Introduction

Neonatal encephalopathy (NE) is one of the most common causes of neonatal morbidity and mortality [1]. The incidence rate of NE varies drastically across regions from 2-30 per 1000 births [2]. Cardiotocography (CTG) is used during labour to detect the fetus at risk of NE and allow for timely intervention. CTG monitors use ultrasound transducers to record fetal heart rate (FHR), maternal heart rate and uterine contractions (UC). The technology was developed in the 1950s to help identify the fetus at risk of oxygen deprivation and has since been widely accepted as the gold standard for fetal monitoring during labour [3].

The International Federation of Gynaecology and Obstetrics (FIGO) guidelines outline features that characterise reassuring and non-reassuring CTG, as outlined in Table 1 [4]. Excessively high or low FHR baseline and/or variability are indicative of non-reassuring CTG. FHR decelerations (decels) and accelerations (accels) in response to contractions are healthy and are an indication of a responsive fetus. However, excessive, prolonged or late decels in response to contractions indicate that the fetus is not coping well and may be suffering a hypoxic event, whereby oxygen supply is restricted [4]. The medical team interpret the CTG in real time during labour and are vigilant for such events. However, inter- and intra-observer disagreement in classifying such events is high [3]. Therefore, the effectiveness of CTG support is widely debated. Incorrect CTG interpretation may have severe implications, including stillbirth, neonatal death and brain injury.

A recent systematic review examined the effectiveness of AI-based CTG interpretation systems that had been evaluated in randomized clinical trials (RCTs) [5]. The review concluded that these AI systems did not improve patient outcomes. It found that agreement between the AI systems and humans was moderate, but the AI systems used aimed to mimic human interpretation, which was akin to adding a "second evaluator with similar instructions" [6] [7]. It suggests that an effective CTG decision-support tool needs to add value to the diagnostic process by using features that are not already interpreted and assessed by human eye.

Several novel feature engineering techniques have been used in prior art to develop CTG features for predicting fetal compromise with promising results [8] [9]. As an alternative to conventional measures of signal variability, phase-rectified signal averaging has been used to compute the mean decelerative capacity of FHR [10]. Retrospective analysis showed that decelerative capacity provides higher predictive value in comparison to short-term variability, which is widely used in clinical practice for the task of predicting fetal acidosis. The use of fractal analysis and the Hurst parameter has been shown to provide a robust alternative to arbitrarily defining frequency bands and computing spectral density [11]. CTG traces are inter-related signals, meaning that the FHR and UC traces can be viewed as a system, as opposed to independent signals. Promising results were achieved by modelling the dynamic relationship between FHR and UC as an impulse response function [12]. This approach is clinically relevant to identify normal versus abnormal decelerations. Deep learning methods have been proposed, whereby the algorithm itself defines what patterns are discriminant. AI architectures, based on convolutional neural networks (CNN) trained on over 35,000 patients were recently published [13]. The CNN used FHR, UC and a signal quality measure as the input data and showed favourable performance compared to clinical practice in the retrospective analysis. However, deep learning models require vast amounts of data and high signal quality to achieve suitable performance.

This paper presents the design of novel features that characterise the response of the FHR to contractions using an ARMA model. Machine learning (ML) models are trained using the ARMA features to predict fetal compromise. The effect of signal quality on the ARMA features is evaluated in the machine learning model. Standard CTG features and clinical variables from the electronic health records (EHR) are subsequently added to the feature set to assess improvements in overall classifier performance. Adding clinical variables provides better clinical context and is a vital part of the diagnostic process and decisions to intervene [14].

* Authors contributed equally.
[1] Infant Research Centre, University College Cork (UCC), Ireland.
[2] Department of Obstetrics and Gynaecology, UCC, Cork, Ireland.
[3] School of Engineering, UCC, Ireland.
[4] Department of Paediatrics & Child Health UCC, Ireland
Research was supported by Science Foundation Ireland (19/FIP/AI/7483).

TABLE I. DEFINITIONS OF REASSURING AND NON-REASSURING CTG

|  | *Reassuring* | *Non-reassuring* |
|---|---|---|
| baseline | 110-160 bpm | < 110 bpm or > 160 bpm |
| variability | 5-25 bpm | < 5 bpm or > 25 bpm |
| decels | early decels: Short and coinciding with UC | late decels: > 20 secs. after UC<br>Prolonged decels: > 3 mins.<br>Repetitive decels: occurring with > 50% of UCs |

## II. METHODOLOGY

### A. Data

The Czech Technical University and University Hospital Brno (CTU-CHB) collected CTGs from 552 patients, which were made publicly available on the PhysioNet platform [15]. The CTGs were recorded within 90 minutes of birth with 60 minutes of Stage I labour and up to 30 minutes of Stage II labour. The dataset contained singleton, full-term pregnancies without prior-known developmental issues. The sampling rate of the CTG was 4 Hz, and it included the FHR and UC traces. The CTU-CHB database has a limited number of clinical variables, including maternal age, parity, gravidity, gestation, limited medical history, delivery type and the duration of labour stages.

It includes cord pH and Apgar scores, which are used clinically to assess neonatal wellbeing after birth. A pH of less than 7.05 and low Apgars, which is a subjective measure from 0 to 10, indicates signs of asphyxia [16]. pH as a single proxy is used in prior art to create labels for classifying the healthy fetus versus the compromised fetus [10] [11]. In this study, a composite label was used, which uses both pH and Apgar (at 5 minutes after birth) to label patients as 'healthy' or 'at-risk'.

### B. Pre-processing

CTG signals are often noisy, containing outliers, artifacts and gaps in traces. In this study, software tools were developed for automated signal pre-processing and cleaning. A feature was also created to assess the ratio of continuous trace versus missing samples. Patients with over 30% of traces missing were removed from the study. FHR samples were deemed to be artifacts based on the following thresholds and criteria:

- Outliers below 50 bpm and above 210 bpm.
- Change greater than 30% from moving average.
- Removal of MHR by discarding sequences of significantly lower FHR values than the baseline between abrupt signal changes.
- Gaps in samples.

### C. Feature engineering

Over 140 standard features from stats, time, frequency and non-linear domains were computed based on prior art, such as baseline, variability, entropy, wavelets and spectral densities [9]. A peak detection algorithm was developed to annotate peaks longer than 10 seconds and with a prominence of > 20% above or below the baseline. Figure 1 shows an example of a CTG with automatically detected contractions, accels and decels. Features were subsequently computed, including the mean and max durations, heights, prominences and number of events, and the ratio of events to background CTG.

The differentiator between normal and abnormal decels is the time from contraction to decel (early decels vs. late decels) [4]. Therefore, a control theory approach is proposed to model the CTG as a system, whereby the UC trace is the excitation signal and the FHR is the output signal of the system.

The following linear ARMA model was used to capture this dynamic interaction [17]:

$$FHR(k) = \sum_{i=1}^{n} \hat{\alpha}_i FHR(k-i) + \sum_{j=1}^{m} \hat{\beta}_j UC(k-j) + \varepsilon(k)$$

where $\varepsilon$ represents unmodelled disturbances and noise.

A windowed least squares algorithm was used to estimate the model parameter vector for each window of 5000 samples. For the $p^{th}$ window, the parameter vector ($\hat{\theta}(p)$):

$$\hat{\theta}(p) = \left[ \hat{\alpha}_1(p), \ldots \hat{\alpha}_n(p), \hat{\beta}_1(p), \ldots \hat{\beta}_m(p) \right]^T \text{is identified using}$$

$$\hat{\theta}(p) = \left( \Phi(p)^T \Phi(p) \right)^{-1} \Phi(p)^T Y(p)$$

where $\Phi(p)$ is the regressor matrix constructed from the input UC and output FHR samples and $Y(p)$ is the target vector constructed from the FHR samples from the $p^{th}$ window.

Changes in the dynamic model over the duration of the CTG can be detected by observing the distribution of the model poles over all of the models generated by the windowed least squares algorithm over the trace. The poles of the model extracted from the $p^{th}$ window are the roots of the characteristic equation:

$$\{r_1(p) \ldots r_n(p)\} = roots\left(z^n + \hat{\alpha}_1(p)z^{n-1} + \hat{\alpha}_2(p)z^{n-2} + \hat{\alpha}_n(p)\right)$$

The magnitude of each pole was used as a measure of its speed. The measure of change over the constituent windows of CTG for the 1st and 2nd poles ($\Delta R_1$, $\Delta R_2$) was:

$$\Delta R_i = \max_p |r_i(p)| - \min_p |r_i(p)|$$

The features $\Delta R_1$ and $\Delta R_2$ were added to the feature set to capture the relationship between the UC and the FHR, and to differentiate between early and late decels.

### D. Machine Learning Models

Binary classification was used to detect fetal compromise. The patients were grouped into *normal* and *at-risk* cases based on two proxy metrics: pH and Apgar at 5 minutes (Apgar 5). Patients that did not fit into either class (n=99) were discarded at this stage of analysis. The labels were defined as:

- Normal (n=310): pH >= 7.15 and Apgar 5 >= 9
- At-risk (n=23): pH <= 7.0 or Apgar 5 <= 6

ML models were evaluated using several metrics including AUC, true positive rate (TPR) and false positive rate (FPR). Leave-One-Out (LOO) cross validation (CV) was used to give a robust prediction of the patient-independent generalisation performance of the classifier. Here, all the data from one patient is removed and the classifier is then trained on the data from the remaining patients, using a stratified 5-fold CV for internal model selection. For each training set, this therefore yields 5 trained models. The performance of an ensemble of these 5 models is then determined over the held out test patient. This process is repeated until each patient in turn has been used once as the test patient. The LOO performance is then reported as the mean performance over all test patients.

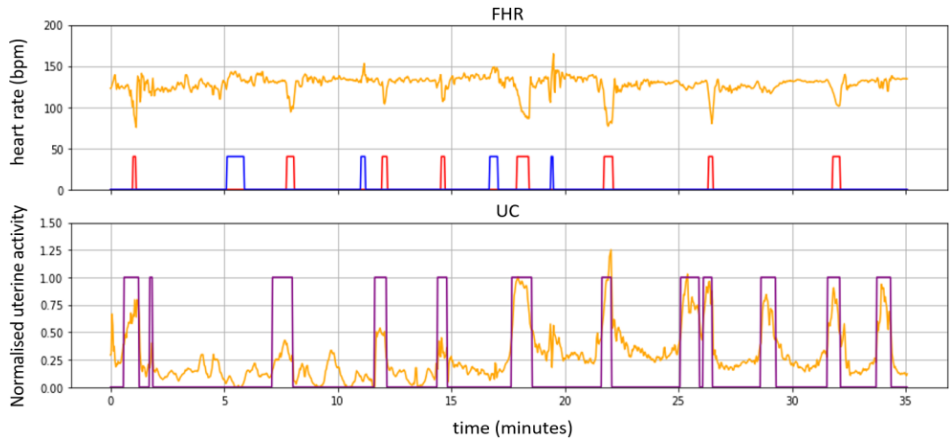

Figure 1. CTG trace after pre-processing and automated event detection (red: decelerations, blue: accelerations and purple: contractions)

Models were trained using stratified 5-fold Cross Validation (CV). 80% of patients were used for training and 20% for validation. Model performance was averaged across the 5 folds. To assess the generalisation error of the models, Leave-One-Out (LOO) CV was used, whereby one patient was taken out and 5 models were trained on the rest of the patients using 5-fold CV.

Prediction on the left-out patient was then made by 5 models. To average predictions, two ensemble methods were used: average probabilities and majority voting. In the first method, the probabilities obtained by the 5 individual CV models were averaged and a 0.5 threshold was applied to decide the results. In majority voting, the binarized predictions were obtained from each model and then those predictions were voted by the 5 models to produce the final result. Both approaches produced similar results, with averaging probabilities being slightly better. The procedure was repeated for all patients in the loop, so that each patient was in turn will have been the test patient. The results were then accumulated across all patients.

*E. Feature Selection*

A robust feature selection routine was implemented to reduce the number of features and select an optimal feature set. The following methods were used:

1. Calculated Pearson Correlation between features and removed redundant features with correlation > 88%.
2. Added features one-by-one and optimised MCC on 5-folds CV. MCC allows for optimisation of all quadrants of confusion matrix [18]. This provided an approximation of features with predictive value.
3. Lastly, features were manually selected and fine-tuned by optimizing MCC averaged in a 5-fold CV.

III. RESULTS

Assessment of the ARMA features with n=2 and m=1 was used to visualise the decision boundary. An SVM was trained on 35 patients that were verified by an obstetrician to be high quality and correctly labelled. Using ARMA features only ($\Delta R_1$ and $\Delta R_2$), Figure 2 shows the decision boundary created between the two classes (blue: normal, red: at-risk). The trained SVM was tested on the remaining patients to assess its performance.

TABLE II. SVM RESULTS ON DIFFERENT HOLD-OUT TEST-SETS

| Test set | AUC | TPR | FPR |
|---|---|---|---|
| all held-out patients | 0.63 | 0.57 | 0.30 |
| held-out patients with 75% intact | 0.67 | 0.62 | 0.29 |
| held-out patients with 80% intact | 0.70 | 0.64 | 0.20 |

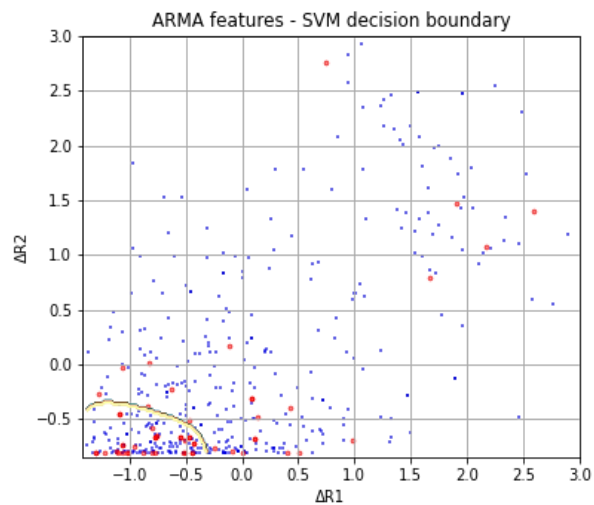

Figure 2. SVM decision boundary using the ARMA features after normalisation (blue: normal, red: at-risk)

Prior to testing, a measure of FHR signal quality (ratio of missing duration to full duration in the trace) was computed for each patient. Based on this, 3 test-sets were created: all held-out patients, held-out patients with greater than 75% of signals intact, and held-out patients with greater than 80% of signals intact. The AUC, TPR and FPR for each test set is shown in Table II. It is evident that performance is better on higher-quality data. Similar trends were reported in prior studies for deep learning models [13].

The ARMA features were then included in the wider CTG feature set of over 140 features, described in Section II(C). Logistic regression models performed best on the wider feature sets in preliminary training. The logistic regression models were trained using 5-folds CV and tested via LOO. The different feature sets (FS) were selected using the feature selection routine described in Section II(E):

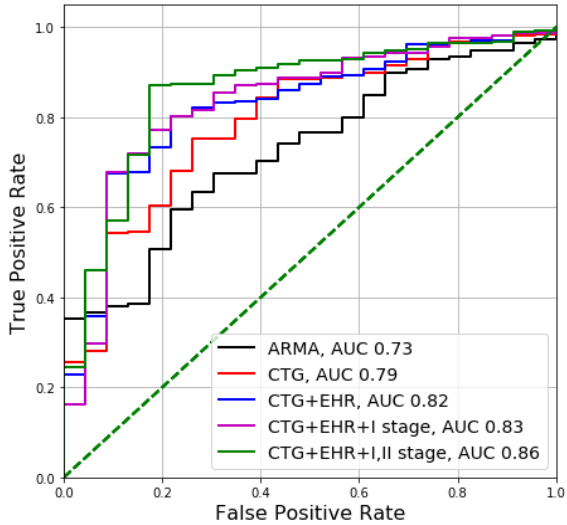

Figure 3. Logistic regression AUC curves for the different feature sets

TABLE III. LOGISTIC REGRESSION RESULTS ON DIFFERENT FEATURE SETS USING 5-FOLDS CV AND LOO

| features set | 5-folds CV | | | Leave-one-out | | |
|---|---|---|---|---|---|---|
| | AUC | TPR | FPR | AUC | TPR | FPR |
| CTG features | 0.735 | 73.9 | 29.0 | 0.79 | 73.9 | 29.3 |
| CTG + EHRs | 0.793 | 82.6 | 25.5 | 0.825 | 82.6 | 26.8 |
| CTG + EHRs + I stage duration | 0.794 | 82.6 | 25.1 | 0.83 | 82.6 | 25.5 |
| CTG + EHRs + I & II stage dur. | 0.809 | 82.6 | 22.3 | 0.86 | 82.6 | 22.2 |

*FS1 (CTG)*: FHR range, max, median, and autocorrelation50 (using the open-source *tsfresh* library [19]), contraction mean prominence and the ARMA features ($\Delta R_1$ and $\Delta R_2$).
*FS2 (CTG+EHR)*: FS1 plus mother's parity, gestation and history of hypertension. An AUC of 0.79 was achieved using CTG features only.
*FS3 (CTG+EHR+Istage)*: FS2 plus Stage I labour duration.
*FS4 (CTG+EHR+I,IIstage)*: FS3 plus Stage II duration.

Figure 3 presents the AUC curves obtained for LOO patients using logistic regression models for the different feature sets. The ARMA features alone achieved 0.73 AUC. The wider CTG feature set achieved 0.79 AUC. Including the clinical variables from the EHR helped improve performance, achieving 0.82 AUC. Including the duration of Stage I and Stage II labour as input features further improved the model's performance to 0.83 AUC and 0.86 AUC, respectively.

In addition, TPR and FPR metrics were calculated to compare against current clinical practice. Prior studies estimated that clinicians currently achieve 31% TPR at 15% FPR in detecting fetal compromise using CTG [13]. The TPR and FPR results for the logistic regression models in 5 fold CV and LOO are summarised in Table III. With the proposed CTG features alone, the machine learning model detects 74% of at-risk cases, although at a higher FPR of 29%. Combining CTG features and clinical factors from EHRs improves the model, achieving 82.6% TPR at 27% FPR.

These results show that combination of EHRs with features extracted from CTG provide better performance for detection of fetal compromise. Adding the duration of Stage I and II labour to the classifier further improves performance of the logistic regression model. Including both duration of Stage I labour and Stage II labour as input features results in 82.6% TPR at a 22% FPR.

## IV. DISCUSSION

Figure 2 demonstrates that the ARMA features provide valuable information to the classifier. The decision boundary created by the trained SVM separates the classes with a relative degree of accuracy. Testing the SVM on unseen data, as in Table II, shows the dependence of the ARMA features on signal quality. Performance increases from 0.63 AUC to 0.70 AUC when tested on unseen patients with higher signal quality. The ARMA features are dependent on both the FHR and UC trace to determine the response time. In general, as the patient nears delivery, maintaining a high-quality UC trace can become clinically difficult, which adds a significant challenge.

In recent years, the use of EHRs in maternity hospitals has grown. The patient information available in these EHRs are vital to the diagnostic process. The results in this study show that performance is enhanced by including clinical variables in the classifier, including gestation, parity and hypertension. All these clinical variables are known prior to labour and can be used to inform decision making. The CTU-CHB database, however, has a very limited set of clinical variables. Based on the results presented in this paper, it is proposed that using a wider range of clinical variables, such as those available in modern EHRs will significantly improve the performance in classification of fetal compromise during labour, and will help to create an unbiased risk assessment for each patient.

The durations of Stage I and Stage II labour were included in the final feature sets. These were separately held out as they may not be known prior to labour to inform decision making. However, an approach could be used whereby the duration of Stage II labour is included as the labour progresses for a real-time decision support system with evolving risk.

In comparison to prior art that used the same open-source database, the results in this paper perform favourably. Prior art on the dataset achieved 64% TPR at 35% FPR [9], 72% TPR at 35% FPR [20], and 40% TPR at 14% FPR [21].

The main limitation of the study is the size of the database. In order to develop a robust detection algorithm, hold-out testing is required to assess the generalisability of the developed features on unseen data. The publicly available database was not of sufficient size to both train the model and have a hold-out test set. To mitigate this, two validation methods were used, 5-fold CV and LOO. The results obtained using both validation methods were similar, which indicates that the model performance is reliable for this type of data.

Future work in this field will include the testing of the developed machine learning models on a larger database of CTG recordings, with a wider range of clinical variables. This paper successfully demonstrates the feasibility of using ARMA models to capture the dynamic relationship between the UC and FHR traces, which bears significant physiological importance, as per the FIGO guidelines. Further work in optimising the window length and order of the ARMA model will be conducted. Finally, the end goal is to deploy this system as a real-time decision support tool. Next steps will use overlapping windows of 30 minutes. Features will be computed on each epoch and the classifier will be re-trained. Predictions could then be made in real-time, providing the medical team with objective decision support.

## V. Conclusion

This study presents novel signal processing and feature extraction methods for the application of CTG classification. Features from the control-theory domain offer promising results for accurate classification of fetal compromise through modelling the FHR time response in relation to contractions. The features are highly dependent on signal quality however. Using a combined CTG feature set that includes contraction detections and statistical features provides improved results. Similarly, the inclusion of clinical risk factors leads to improved classifier performance and may be vital for future clinical adoption of the technology. In all cases, the results out-perform current clinical practice and offer a promising method for automated detection of fetal compromise.

## VI. References

[1] A. C. Lee, N. Kozuki, H. Blencowe, T. Vos, A. Bahalim, G. L. Darmstadt, S. Niermeyer, M. Ellis, N. J. Robertson, S. Cousens and J. E. Lawn, "Intrapartum-related neonatal encephalopathy incidence and impairment at regional and global levels for 2010 with trends from 1990," *Pediatric Research,* vol. 74, no. 1, pp. 50-72, 2013.

[2] H. Namusoke, M. M. Nannyonga, R. Ssebunya, V. K. Nakibuuka and E. Mworozi, "Incidence and short term outcomes of neonates with hypoxic ischemic encephalopathy in a Peri Urban teaching hospital, Uganda: a prospective cohort study," *Maternal Health, Neonatology and Perinatology,* vol. 4, no. 6, pp. 0-6, 2018.

[3] A. Pinas and E. Chandraharan, "Continuous cardiotocography during labour: Analysis, classification and management," *Best Practice & Research Clinical Obstetrics & Gynaecology,* vol. 30, pp. 33-47, 2016.

[4] D. Ayres-de-Campos, C. Y. Spong and E. Chandraharan, "FIGO consensus guidelines on intrapartum fetal monitoring: Cardiotocography," *International Journal of Gynaecology and Obstetrics,* vol. 131, no. 1, pp. 13-24, 2015.

[5] J. Balayla and G. Shrem, "Use of artificial intelligence (AI) in the interpretation of intrapartum fetal heart rate (FHR) tracings: a systematic review and meta-analysis," *Archives of Gynecology and Obstetrics,* vol. 300, pp. 7-14, 2019.

[6] R. D. F. Keith and K. R. Greene, "Development, evaluation and validation of an intelligent system for the management of labour," *Baillière's Clinical Obstetrics and Gynaecology,* vol. 8, no. 3, pp. 583-605, 1994.

[7] D. Ayres-de-Campos, P. Sousa, A. Costa and J. Bernardes, "Omniview-SisPorto 3.5 - a central fetal monitoring station with online alerts based on computerized cardiotocogram+ST event analysis," *Journal of Perinatal Medicine,* vol. 36, no. 3, pp. 260-264, 2008.

[8] J. Spilka, J. Frecon, R. Leonarduzzi, N. Pustelnik, P. Abry and M. Doret, "Sparse Support Vector Machine for Intrapartum Fetal Heart Rate Classification," *IEEE Journal of Biomedical and Health Informatics,* vol. 21, no. 3, pp. 664-671, 2017.

[9] J. Spilka, G. Georgoulas, P. Karvelis, V. P. Oikonomou, V. Chudácek, C. Stylios, L. Lhotská and P. Janku, "Automatic Evaluation of FHR Recordings from CTU-UHB CTG Database," *International Conference on Information Technology in Bio- and Medical Informatics,* pp. 47-61, 2013.

[10] A. Georgieva, A. T. Papageorghiou, S. J. Payne, M. Moulden and C. W. Redman, "Phase-rectified signal averaging for intrapartum electronic fetal heart rate monitoring is related to acidaemia at birth," *BJOG An International Journal of Obstetrics and Gynaecology,* vol. 121, no. 7, pp. 889-894, 2014.

[11] M. Doret, J. Spilka, V. Chudácek, P. Goncalves and P. Abry, "Fractal analysis and hurst parameter for intrapartum fetal heart rate variability analysis: a versatile alternative to frequency bands and LF/HF ratio," *PloS One,* vol. 10, no. 8, 2015.

[12] P. A. Warrick, E. F. Hamilton, D. Precup and R. E. Kearney, "Classification of Normal and Hypoxic Fetuses From Systems Modeling of Intrapartum Cardiotocography," *IEEE Transactions on Biomedical Engineering,* vol. 57, no. 4, pp. 771-779, 2010.

[13] A. Petrozziello, C. W. Redman, A. T. Papageorghiou, I. Jordanov and A. Georgieva, "Multimodal Convolutional Neural Networks to Detect Fetal Compromise During Labor and Delivery," *IEEE Access,* vol. 7, pp. 112026-112036, 2019.

[14] A. Georgieva, C. W. Redman and A. T. Papageorghiou, "Computerized data-driven interpretation of the intrapartum cardiotocogram: a cohort study," *Acta Obstetricia et Gynecologica Scandinavica,* vol. 96, no. 7, pp. 883-891, 2017.

[15] V. Chudácek, J. Spilka, M. Bursa, P. Janku, L. Hruban, M. Huptych and L. Lhotská, "Open access intrapartum CTG database," *BMC Pregnancy and Childbirth,* vol. 14, no. 16, pp. 0-12, 2014.

[16] G. L. Malin, R. Morris and K. S. Khan, "Strength of association between umbilical cord pH and perinatal and long term outcomes: systematic review and meta-analysis," *BMJ,* vol. 340, 2010.

[17] B. K. Nelson, "Time Series Analysis Using Autoregressive Integrated Moving Average (ARIMA) Models," *Academic Emergency Medicine,* vol. 5, no. 7, pp. 739-744, 1998.

[18] D. Chicco and G. Jurman, "The advantages of the Matthews correlation coefficient (MCC) over F1 score and accuracy in binary classification evaluation," *BMC Genomics,* vol. 21, no. 6, pp. 0-13, 2020.

[19] M. Christ, N. Braun and J. Neuffer, "tsfresh," [Online]. Available: https://tsfresh.readthedocs.io/en/latest/index.html.

[20] G. Georgoulas, P. Karvelis, J. Spilka, V. Chudácek, C. D. Stylios and L. Lhotská, "Investigating pH based evaluation of fetal heart rate (FHR)," *Health and Technology,* vol. 7, pp. 241-254, 2017.

[21] J. Spilka, V. Chudácek, M. Huptych, R. Leonarduzzi, P. Abry and M. Doret, "Intrapartum Fetal Heart Rate Classification: Cross-Database Evaluation," *Mediterranean Conference on Medical and Biological Engineering and Computing,* pp. 1199-1204, 2016.